\documentclass[runningheads]{llncs}
\usepackage{graphicx}
\usepackage{hyperref}
\usepackage{multirow}
\usepackage{subfigure}
\usepackage{longtable}

\makeatletter
\newcommand{\printfnsymbol}[1]{%
  \textsuperscript{\@fnsymbol{#1}}%
}
\makeatother
\begin{document}
\title{Context-aware Retail Product Recommendation with Regularized Gradient Boosting}
%
%
\author{Sourya Dipta Das\thanks{equal contribution}\inst{1} \and
Ayan Basak\printfnsymbol{1}\inst{1}}
\authorrunning{Sourya et al.}
%
\institute{Razorthink Inc,USA \\
\email{\{dipta.juetce,ayanbasak13\}@gmail.com}}

\titlerunning{Context-aware Retail Product Recommendation}
\maketitle              
\begin{abstract}
In the FARFETCH Fashion Recommendation challenge, the participants needed to predict the order in which various products would be shown to a user in a recommendation impression. The data was provided in two phases - a validation phase and a test phase. The validation phase had a labelled training set that contained a binary column indicating whether a product has been clicked or not. The dataset comprises over 5,000,000 recommendation events, 450,000 products and 230,000 unique users. It represents real, unbiased, but anonymised, interactions of actual users of the FARFETCH platform.
The final evaluation was done according to the performance in the second phase. A total of 167 participants participated in the challenge, and we secured the 6th rank during the final evaluation with an MRR of 0.4658 on the test set. We have designed a unique context-aware system that takes the similarity of a product to the user context into account to rank products more effectively. Post evaluation, we have been able to fine-tune our approach with an MRR of 0.4784 on the test set, which would have placed us at the 3rd position.

\end{abstract}

\section{Introduction}
With the advancement of digital media, the online retail sector has started to gain a lot of importance. Coupled with other factors like the COVID-19 pandemic, people often tend to prefer online shopping for fashion items. Hence, the content that is shown to users on fashion websites is very important. In order to sustain user interest on a platform, it is important to display as much relevant information as possible and also rank the items according to their relevance. Hence, the development of a robust recommendation system is very important. The main task in the FARFETCH Recommendation Challenge is to predict the optimal ranking of the 6 products in a given impression, such that products with higher probability of being clicked by the user in that recommendation context have lower rank (first positions). We can use click through data and product attributes along with the contextual similarity of a product in order to improve the quality of recommendations. We have tried to approach the problem using various ranking algorithms as well as a binary classification task, and the latter approach has proven to be more effective. We have been able to improve our results by around 15\% using a context-aware post-processing approach that makes use of the similarity of a product to the user context, achieving a final MRR of 0.4814 on the validation set and 0.4658 on the test set during the original course of the competition. Post evaluation, we have been able to improve our performance on the test set drastically by fine-tuning the threshold parameter in our post-processing approach, achieving an MRR of 0.4784.

\section{Problem Description}
The problem statement has been provided as a part of the FARFETCH Fashion Recommendations Challenge 2021~\cite{farfetch2021challenge}. We have been provided with a large sample of FARFETCH's recommendations system impressions and associated click events. An impression is a list of 6 products that were shown to a user in a recommendations UI module in a given context(type of page, date, etc), with a label identifying which of the 6 products have been clicked. The goal of the competition is to predict the optimal ranking of the 6 products in a given impression, such that products with a higher probability of being clicked by the user in the recommendation context has lower rank.
\subsubsection{Dataset Description : }
The dataset is a large sample of FARFETCH's recommendations system impressions and associated click events, captured over a period of 2 months. Each row of data has a query\_id which is a unique identifier of a recommendation impression. In the initial phase of the competition, we had access to a training set of 3.5 million rows that includes a target label, is\_click, and a validation set of about 0.6 million rows without a target label for making the initial submissions. The dataset contains 229064 unique users, 443150 unique products, and 813729 unique impression lists. 
A separate attribute dataset is provided which contains the attribute information for each of the products. Considering all the data present, there are 3 numerical attributes and around 20 categorical ones.
The target label identifies which of the 6 products in the impression were clicked. The position of the products within the list is unknown in this dataset.
In the final phase, we had access to another unlabelled test set with around 0.6 million rows for final evaluation. 
\subsubsection{Evaluation Metric : }
During the competition, models were evaluated using Mean Reciprocal Rank (MRR), defined as follows:
\begin{equation}
MRR=\frac{1}{|N|} \sum_{i=1}^{|N|} \frac{1}{rank_{i}}
\end{equation}
where, where N is the number of impressions, and for the i-th impression in the dataset, $rank_i$ is the rank of the first correct prediction.

\section{Methodology} 
Initially, we tried to approach the problem as a ranking task with lambdarank as the objective function. However, further experimentation proved that approaching it as a simple binary classification problem to predict the clickout probability of a product using the binary cross-entropy objective, and ranking the products using the prediction probabilities yield much superior results. One of the most important steps in our solution was the feature engineering part, since it was challenging as well as time-consuming. We have created around 80 features based on user, session, product, clickout, and other relevant attributes. Our results have been further boosted using the context-aware post processing technique that we have employed.

\subsection{Preprocessing}
The main data preprocessing that we needed to employ in this problem was categorical variable encoding. The given dataset has a lot of categorical columns, most of which have a large number of distinct categories. We have used label encoding as the standard categorical variable encoding technique over approaches like one-hot encoding due to the sparsity in the data. 

\subsection{Feature Extraction}
This is the most important step of our solution. A carefully engineered feature may have the potential to explain the significance of a product within a query quite effectively. Broadly, we have two types of features- click-out features and non click-out features.

\subsubsection{Non Click-out Features} \hfill \break 
\hfill \break 
These features are not dependent on whether a particular product has been clicked out or not. They are further sub-divided into the following categories: \par
\textbf{Session-based Features: } 
These features include session specific aggregate and frequency features. Aggregate features include the mean, maximum, minimum product price and start online date in a session while frequency features include product frequency, main colour frequency, etc in a particular session. For example, the \textit{product\_session\_click\_proportion} feature, which is the most important feature of our best performing method, is a type of repeated clicks rate feature. It is the ratio of the total number of clicks on the product under consideration in a particular session to the total number of clicks on all the products in the same session. 

\textbf{Query-based Features}: 
These features include the aggregated like mean, maximum and minimum product price and start online date within a particular query.

\textbf{Global Static Features}:
Features like the total number of users who have interacted with a particular product, number of impressions that a particular product appears in, etc. compose this set of features.

\textbf{Popularity Features}:
These features provide information about various product attributes like popularity of brand id, category id, main colour, etc.

\textbf{User-tier features}:
User-tier specific aggregate features like the mean, maximum and minimum product price and start online date are included in this sub-category of features.

\textbf{Ranking Features}:
This features are basically the ranking of a product within an impression based on the product price and start online date.

\textbf{Difference Features}:
These features basically convey the comparison of a particular feature against an aggregate like feature, for example, the difference between a particular product price and the mean product price in that session.

\subsubsection{Click-out Features} \hfill \break 
\hfill \break 
These features are dependent on whether a particular product has been clicked out or not. They are further sub-divided into the following categories:\par
\textbf{Global Static features}:
These features include details about click interactions of a product like number of unique users who have clicked on a particular product, the price of a product when it was last clicked, etc.

\textbf{Popularity features}:
These features indicate the popularity of an attribute by computing percentage of click on a particular attribute. For example, the percentage click on a particular brand id among all clicked brands, the percentage click on a particular main colour among all clicked main colours, etc.

\textbf{User-tier features}:
Users are segmented into various tiers based on some attributes. The user-tier features include the aggregates like mean, maximum and minimum product price and start online date of clicked products based on the user tier.

\textbf{Difference Features}:
The difference features include the difference of product price and start online date and the mean of those features among the clicked products. They convey how a particular product attribute compares to the same attribute among clicked products.

\textbf{Weekly features}:
These features explain attribute variation at a weekly level. For example, the mean price of clicked products each week, the frequency of product clicks each week, etc. make up some of these features.

\subsubsection{Product-Context Similarity(PCS) Feature}\hfill \break 
\hfill \break 
The main motivation behind incorporation context-awareness in the recommendation system is that it enables us to generate recommendations more relevant to the specific contextual situation of an user. A context-based recommendation system gains special importance when information from multiple contexts are aggregated. For example, the products that a user might be interested in from the “Books” context might be totally irrelevant and have no predictive value for recommendations for the same user when he decides to purchase some products from the “Clothing” context. Gediminas et al.~\cite{adomavicius2011context} has shown how context provides additional capabilities in recommendation systems and how the general notion of context can be modelled in recommendation systems.

We have designed a PCS feature, that computes the similarity of a particular product to its corresponding user context. The context is basically a product that a user originally searched for, and recommendations are shown based on the context item. Hence, the similarity of the context item to another product provides an understanding about the relevance of a particular product to the original context. The PCS value is computed as follows:

For each product $P_i$, let  $U^{i}_{c} = \{c^{1}_{i},c^{2}_{i}, \dots c^{k_{1}}_{i}\}, U^{i}_{f} = \{f^{1}_{i},f^{2}_{i}, \dots f^{k_{2}}_{i}\}, U^{i}_{l} = \{l^{1}_{i},l^{2}_{i}, \dots l^{k_{3}}_{i}\}$, where $U^{i}_{c}$ is the set of single-element categorical features, $U^{i}_{f}$ is the set of numerical features, and $U^{i}_{l}$ is the set of multiple-element (list) categorical features.

$S(P_i,P_j) = \frac{1}{N_p}[\sum_{h=0}^{k_{1}}\phi_C(c^{h}_{i}, c^{h}_{j}) + \sum_{h=0}^{k_{3}}\frac{n(l^{h}_{i} \cap l^{h}_{j})}{max(n(l^{h}_{i}), n(l^{h}_{j}))} + \sum_{h=0}^{k_{2}}(1 - |f^{h}_{i} - f^{h}_{j}|)]$
\\
where, $S(P_i,P_j)$ is the PCS value between $P_i$ and $P_j$, $N_p$ is the total number of product attributes,
$\phi_C(c^{h}_{i}, c^{h}_{j})=1 \:if\: c^{h}_{i}=c^{h}_{j}, \:else \:0$,
$l^{h}_{i} \cap l^{h}_{j}$ denotes the common features of the feature lists, and n(.) is the set cardinality.

\subsection{Gradient Boosting Model}
Gradient Boosting is a classical machine learning approach of converting weak learners into a strong one, by gradually, additively, and sequentially training numerous models. In the boosting approach, a new tree is a fit on a modified version of the original data set. We have employed this approach to build our classifier model. 

\subsubsection{Ranking Approach}
We have used LightGBM for the ranking approach. In order to use it for ranking, we use lambdarank as an objective function. The idea of lambdarank is to use the gradient of cost with respect to model score instead of cost. An important input to the ranking model is a list containing the length of each group, or the number of products in each impression, i.e. number of impressions corresponding to each query id, besides the features and the labels. The products within a query are ranked using the predicted probability values obtained using the trained LightGBM Ranker model.

\subsubsection{Binary Classification Approach}
In this approach, we treat the task as a simple binary classification problem with the target being whether a product is clicked or not, i.e. is\_click is 0 or 1. A value is predicted between 0 to 1, which indicates the probability of the product being clicked, where a value closer to 1 denotes a higher click probability. We have experimented with LightGBM-\cite{ke2017lightgbm} and XGBOOST~\cite{chen2016xgboost} classifiers. The latter one has proven to be more effective as, we can notice from the results Table \ref{result_table} that the XGBOOST model achieves a higher MRR than the LightGBM model in both the cases, with and without filtering on both validation and test sets. The model is trained using binary cross-entropy as the objective function. During inference, we rank the products within a query using their predicted probability values obtained from the model, i.e. the probability that the product will be clicked.

\subsection{Post Processing}
In order to improve the ranking of products within an impression, we have employed context-aware post-processing technique. Basically, we need to decide upon a threshold product-context similarity value. If the maximum product-context similarity within an impression is lower than this threshold, we rank the products within the impression using the model prediction probabilities. However, if the maximum value is higher than the threshold, we rank the products in the impression using the product-context similarity values only.
The threshold value for a particular dataset is decided upon using an ablation study of the product-context similarity column with different threshold values. Post evaluation, after the availability of the validation and test labels, we have observed that setting a threshold value around 0.45-0.50 helps us achieve the best results.

\section{Experiments}
We have performed all our experiments on a system with 32GB RAM and 2.2 GHz QuadCore Intel Core i7 Processor. It was also equipped with a NVIDIA GTX 1080 GPU with 12 GB VRAM. We have tried to approach this problem both as a ranking task as well as a binary classification problem. The latter approach has proven to be more effective since the XGBOOST model achieves a higher MRR than the LightGBM model in both the cases, with and without filtering on both validation and test sets as shown in Table \ref{result_table}. We have also been able to improve our results drastically using a post-processing approach, making use of the similarity of a product to the corresponding user context.

\subsubsection{Experimental Settings : }
In the case of the ranking approach, we have used the LightGBM-\cite{ke2017lightgbm}Ranker algorithm with lambdarank as the objective function. In the binary classification approach, we have run 2 experiments using LightGBM and XGBOOST~\cite{chen2016xgboost} with binary cross-entropy as the objective function.

\subsubsection{Training and Inference Process : }
We had access to a labelled dataset of around 3.5 million rows. We have randomly sampled out 25000 query ids, i.e. 150,000 rows without replacement and used it as a validation set to tune our models during the competition. The remaining rows in the labelled dataset was used to train our model. Whenever the MRR on the validation set is more than that of a previous iteration, we save the corresponding trained model.
During inference, we load up the best model based on the performance on the validation set and use it to obtain predictions. After obtaining the raw model predictions, we apply the post-processing technique to rank the products within a query.

\subsubsection{Hyperparameter Tuning : }
In each of our approaches, we need to tune around 10 hyperparameters. Some of these include learning rate, max leaves, max depth, etc. We observed that we need to search over a fairly large range of values to get the optimum values of these parameters. Hence, we are using a Bayesian search approach. More specifically, we are using the BayesSearchCV\footnote{https://scikit-optimize.github.io/stable/modules/generated/skopt.BayesSearchCV.html}  utility from scikit-learn\footnote{https://scikit-learn.org/stable/} and the scikit-opt\footnote{https://scikit-optimize.github.io/stable/} library to get the optimum value of parameters. Our best parameters obtained for each of the models are tabulated in section \ref{sec:best_param} from Appendix \ref{sec:appendix}.

\subsection{Experimental Results}
\begin{figure}[!htb]
  \begin{minipage}{0.49\textwidth}
     \centering
     \includegraphics[width=\linewidth]{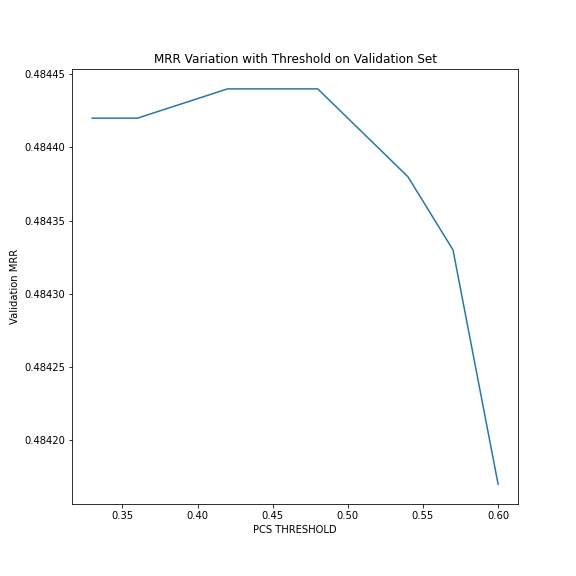}
     \caption{MRR Variation with PCS threshold on Validation Set}\label{Fig:Data1}
  \end{minipage}\hfill
  \begin{minipage}{0.49\textwidth}
     \centering
     \includegraphics[width=\linewidth]{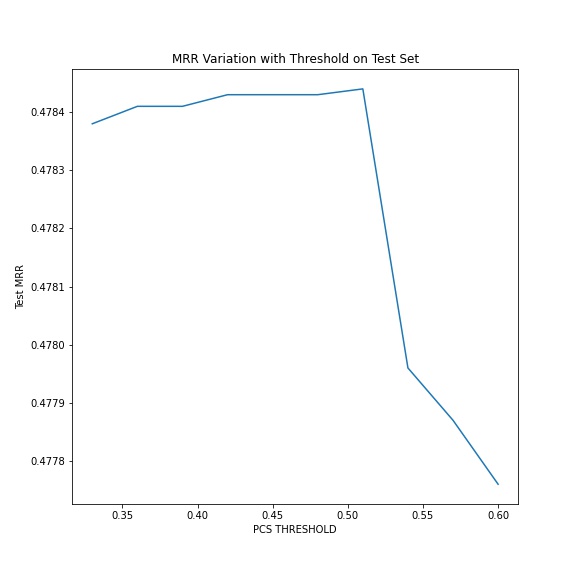}
     \caption{MRR Variation with PCS threshold on Test Set}\label{Fig:Data2}
  \end{minipage}
\end{figure}

The results from our 3 approaches are tabulated in Table \ref{result_table} below. These results have been obtained after fine-tuning the threshold parameter post completion of the competition. As we can observe from the results, the XGBOOST~\cite{chen2016xgboost} Classifier has been able to achieve the best results. The feature importances for this model are tabulated in \ref{sec:featureImp}. We have also been able to improve our final performance on the test set by setting an appropriate threshold value of 0.51. As observed from the results in figures \ref{Fig:Data1} and \ref{Fig:Data2}, the best results are obtained for a threshold value around 0.45-0.50. We can observe that initially, the MRR increases with the increase in the value of threshold till it reaches a peak for both the validation and test sets, and thereafter decreases.
As far as the features are concerned, the proportion of clicks on a product in a particular session seems to be a really important one as can be observed from Table \ref{feature_imp} in Appendix \ref{sec:appendix}. This might be due to the fact that it provides some information about the overall popularity of a product within a user session, and hence, helps the model make a better prediction. The other important features are mainly based on the absolute value of the product price or its related features, like the price of the product when it was clicked last, the difference of a particular product price from the mean price of clicked products, etc. This shows that the price of a product is a major area of concern for users, which is quite expected. The popularity of a product revolves around its price to a great extent.

\begin{table}[]
\small
\centering
\caption{Results of different approaches on validation and test set}
\label{result_table}
\resizebox{\textwidth}{!}{%
\begin{tabular}{|c|c|c|c|c|}
\hline
\multirow{2}{*}{Models} & \multicolumn{2}{c|}{Val} & \multicolumn{2}{c|}{Test} \\ \cline{2-5} 
 & Without filter & With filter & Without filter & With filter \\ \hline
LightGBM-\cite{ke2017lightgbm} Ranker & 0.4088 & 0.4706 & 0.3870 & 0.4712 \\ \hline
LightGBM-\cite{ke2017lightgbm} Classifier & 0.4123 & 0.4840 & 0.3929 & 0.4760 \\ \hline
XGBOOST~\cite{chen2016xgboost} Classifier & 0.4196 & 0.4884 & 0.4111 & 0.4784 \\ \hline
\end{tabular}%
}
\end{table}

\section{Conclusion and Future Work}
In this paper, we have presented an approach to effectively rank the products within a recommendation impression module. We have experimented with both a ranking approach and a binary classification one, and the latter one has proven to be more effective. Our unique product-context similarity feature helps us improve the performance from a vanilla classification approach drastically, and achieve a significantly higher MRR on the validation and test sets. 
In the future, various ensembles of classification techniques or hybrid ensembles of classification and ranking techniques can be tried out in combination with the context-aware post-processing. Also, it might also be worthy to try out deep learning approaches using CNN, LSTM, etc.

\section*{Code}
The code for this paper is available \href{https://github.com/diptamath/Context-aware-Retail-Product-Recommendation}{here}. 
%
%
%
\bibliographystyle{splncs04}
\bibliography{mybibliography}

\begin{thebibliography}{1}
\providecommand{\url}[1]{\texttt{#1}}
\providecommand{\urlprefix}{URL }
\providecommand{\doi}[1]{https://doi.org/#1}

\bibitem{adomavicius2011context}
Adomavicius, G., Tuzhilin, A.: Context-aware recommender systems. In:
  Recommender systems handbook, pp. 217--253. Springer (2011)

\bibitem{chen2016xgboost}
Chen, T., Guestrin, C.: Xgboost: A scalable tree boosting system. In:
  Proceedings of the 22nd acm sigkdd international conference on knowledge
  discovery and data mining. pp. 785--794 (2016)

\bibitem{farfetch2021challenge}
Gonçalves, D., Chaves, I., Gomes, J., Nogueira, P., Otto, T., Marinho, V.:
  Farfetch fashion recommendations challenge --- ecml-pkdd discovery challenge
  2021. Proceedings of ECML-PKDD 2021 Discovery Challenge  (2021)

\bibitem{ke2017lightgbm}
Ke, G., Meng, Q., Finley, T., Wang, T., Chen, W., Ma, W., Ye, Q., Liu, T.Y.:
  Lightgbm: A highly efficient gradient boosting decision tree. Advances in
  neural information processing systems  \textbf{30},  3146--3154 (2017)

\end{thebibliography}

\appendix

\section{Appendix}
\label{sec:appendix}

\subsection{Feature Importance Score}
The table below shows the percentage importance of the top 10 features for the XGBOOST Classifier model.

\label{sec:featureImp}

\begin{table}[]
\small
\centering
\caption{Top 10 Feature Importance for XGBOOST Classifier}
\label{feature_imp}
\resizebox{\textwidth}{!}{%
\begin{tabular}{|c|c|c|}
\hline
\textbf{Feature} & \textbf{\% Importance} & \textbf{Description} \\ \hline
product\_session\_click\_proportion & 39.41 & \begin{tabular}[c]{@{}c@{}}proportion of clicks on a product\\ in a particular session\end{tabular} \\ \hline
\#unique\_users\_clicked & 11.06 & \begin{tabular}[c]{@{}c@{}}\# unique users who have clicked\\ on the product\end{tabular} \\ \hline
last\_clickout\_product\_price & 6.82 & \begin{tabular}[c]{@{}c@{}}product price for the last clickout\\ on the product\end{tabular} \\ \hline
last\_clickout\_days\_elapsed & 5.30 & \begin{tabular}[c]{@{}c@{}}last day (days elapsed) on which\\ clickout was made for a product\end{tabular} \\ \hline
diff\_prod\_price\_from\_click\_mean & 3.33 & \begin{tabular}[c]{@{}c@{}}difference of a product price from \\ the mean price of clicked products\end{tabular} \\ \hline
product\_price & 2.90 & price of the product \\ \hline
days\_elapsed & 2.09 & \begin{tabular}[c]{@{}c@{}}number of days elapsed with respect\\ to the reference date\end{tabular} \\ \hline
start\_online\_date\_max\_clicked\_out\_week\_7 & 1.93 & \begin{tabular}[c]{@{}c@{}}maximum start online date for a\\ clicked product in week 7\end{tabular} \\ \hline
start\_online\_date\_mean\_clicked\_out\_week\_7 & 1.92 & \begin{tabular}[c]{@{}c@{}}mean start online date for a clicked\\ product in week 7\end{tabular} \\ \hline
diff\_prod\_price\_from\_user\_tier\_mean & 1.87 & \begin{tabular}[c]{@{}c@{}}difference of a product price from\\ the mean price of products in \\ current user tier\end{tabular} \\ \hline
\end{tabular}%
}
\end{table}

\subsection{Hyper Parameters of the different Boosting Models}
\label{sec:best_param}

The tables below show the best parameters obtained using Bayesian Parameter tuning for the XGBOOST Classifier, LightGBM Classifier and LightGBM Ranker models, respectively.
\begin{longtable}[c]{|l|l|}
\caption{Best Tuned Parameters for XGBOOST Classifier Model}
\label{tab:xgb_params}\\
\hline
\textbf{Parameter}          & \textbf{Value}  \\ \hline
\endfirsthead
\endhead
learning \_rate    & 0.001  \\ \hline
max\_leaves        & 10     \\ \hline
max\_depth         & 6      \\ \hline
subsample          & 0.71 \\ \hline
colsample\_bytree  & 0.84 \\ \hline
colsample\_bylevel & 0.37 \\ \hline
reg\_alpha         & 7.82 \\ \hline
reg\_lambda        & 5.72   \\ \hline
scale\_pos\_weight & 35     \\ \hline
min\_child\_weight & 6      \\ \hline
booster            & gbtree \\ \hline
\end{longtable}

\begin{longtable}[c]{|l|l|}
\caption{Best Tuned Parameters for LightGBM Classifier Model}
\label{tab:lgb_class_params}\\
\hline
\textbf{Parameter}  & \textbf{Value} \\ \hline
\endfirsthead
\endhead
learning \_rate     & 0.2986         \\ \hline
max\_leaves         & 8              \\ \hline
max\_depth          & 6              \\ \hline
subsample           & 0.90         \\ \hline
bagging\_freq       & 16             \\ \hline
feature\_fraction   & 0.97         \\ \hline
reg\_alpha          & 6.23         \\ \hline
reg\_lambda         & 4.75         \\ \hline
min\_split\_gain    & 0.0338         \\ \hline
min\_child\_weight  & 0.0675         \\ \hline
min\_child\_samples & 21             \\ \hline
n\_estimators       & 8000           \\ \hline
subsample\_for\_bin & 135880         \\ \hline
importance\_type    & split          \\ \hline
boosting\_type      & gbdt           \\ \hline
\end{longtable}

\begin{longtable}[c]{|l|l|}
\caption{Best Tuned Parameters for LightGBM RankerModel}
\label{tab:lgb_rank_params}\\
\hline
\textbf{Parameter}  & \textbf{Value} \\ \hline
\endfirsthead
\endhead
learning \_rate     & 0.01           \\ \hline
max\_leaves         & 15             \\ \hline
max\_depth          & 8              \\ \hline
subsample           & 0.86           \\ \hline
bagging\_freq       & 14             \\ \hline
feature\_fraction   & 0.97           \\ \hline
reg\_alpha          & 6.97           \\ \hline
reg\_lambda         & 7.71           \\ \hline
min\_split\_gain    & 0.0033         \\ \hline
min\_child\_weight  & 0.0469         \\ \hline
min\_child\_samples & 42             \\ \hline
n\_estimators       & 5000           \\ \hline
subsample\_for\_bin & 200000         \\ \hline
importance\_type    & split          \\ \hline
boosting\_type      & gbdt           \\ \hline
\end{longtable}

\end{document}